\documentclass{article} 
\usepackage[accepted]{aistats2016}
\usepackage{url}

\usepackage{graphicx} 

\usepackage{algorithm}
\usepackage{algorithmic}

\usepackage{amsmath, amsthm}
\usepackage{amssymb}
\usepackage{sidecap}  
\usepackage{bm,amsbsy} 

\usepackage[resetlabels,labeled]{multibib}
\newcites{sup}{References}

\usepackage{pifont}
\usepackage{soul}

\makeatletter
\usepackage{todonotes}

\newcommand{\Nrm}{\mathcal{N}}

\newcommand{\Em}{\mathbb{E}}
 
\newcommand{\trp}{{^\top}} 

\renewcommand{\eqref}[1]{Eq.~\ref{eq:#1}}
\newcommand{\figref}[1]{Fig.~\ref{fig:#1}}

\newcommand{\secref}[1]{section~\ref{sec:#1}}

\newcommand{\iid}{\overset{i.i.d.}{\sim}}

\begin{document}

%
\runningtitle{K2ABC}

%
\runningauthor{Park, Jitkrittum, Sejdinovic}

\twocolumn[

\aistatstitle{K2-ABC: Approximate Bayesian Computation \\  with Kernel Embeddings}

\aistatsauthor{ Mijung Park$\phantom{}^{\ast}$ $\phantom{}^{\dagger}$ \And Wittawat Jitkrittum$\phantom{}^{\ast}$ \And  Dino Sejdinovic$\phantom{}^{+}$ }

\aistatsaddress{ $\phantom{}^{\ast}$ Gatsby Unit, University College London, {\{mijungi.p, wittawatj\}@gmail.com} \\ $\phantom{}^{+}$ University of Oxford, {dino.sejdinovic@gmail.com} } ]

\let\oldthefootnote\thefootnote
 \renewcommand{\thefootnote}{\fnsymbol{footnote}} \footnotetext[1]{M Park and W
Jitkrittum contributed equally.} 
 \renewcommand{\thefootnote}{\fnsymbol{footnote}} \footnotetext[2]{Current affiliation: Information institute, University of Amsterdam} 
\let\thefootnote\oldthefootnote

\begin{abstract}
Complicated generative models often result in a situation where computing the likelihood of observed data is intractable, while simulating from the conditional density given a parameter value is relatively easy. 
Approximate Bayesian Computation (ABC) is a paradigm that enables simulation-based posterior inference in such cases by measuring the similarity between simulated and observed data in terms of a chosen set of summary statistics. 
However, there is no general rule to construct sufficient summary statistics for complex models. Insufficient summary statistics will ``leak" information, which leads to ABC algorithms yielding samples from an incorrect (partial) posterior. 
In this paper, we propose a fully nonparametric ABC paradigm which circumvents the need for manually selecting summary statistics. 
Our approach, {\it{K2-ABC}},  uses {\it{maximum mean discrepancy}} (MMD) to construct a dissimilarity measure between the observed and simulated data. The embedding of an empirical distribution of the data into a reproducing kernel Hilbert space plays a role of the summary statistic and is sufficient whenever the corresponding kernels are characteristic. Experiments on a simulated scenario and a real-world biological problem illustrate the effectiveness of the proposed algorithm. 
\end{abstract}

\section{Introduction}


ABC is an approximate Bayesian inference framework for models with intractable likelihoods. 
Originated in population genetics \cite{Beaumont02}, ABC has been widely used in a broad range of scientific applications including ecology, cosmology, and bioinformatics \cite{Ratmann_07, Bazinetal_2010, Schaferetal_12}. 
The goal of ABC is to obtain an approximate posterior distribution over model parameters, which usually correspond to interpretable inherent mechanisms in natural phenomena. However, in many complex models of interest, exact posterior inference is intractable since the likelihood function, the probability of observations $y^*$ for given model parameters $\theta$, is either expensive or impossible to evaluate\footnote{Note that this is different from the intractability of the \emph{marginal likelihood}, which requires integrating out $\theta$ from the likelihood function. While marginal likelihood is often computationally intractable, this issue is readily resolved via a suite of MCMC algorithms.}. This leads to a situation where the posterior density cannot be evaluated even up to a normalisation constant.   


ABC resorts to an approximation of the likelihood function using simulated observations that are {\it{similar}} to the actual observations. 
Most ABC algorithms evaluate the similarity between the simulated and actual observations in terms of a {\it{pre-chosen}} set of summary statistics. Since the full dataset 
is represented in a lower-dimensional space of summary statistics, unless the selected summary statistic $s^*$ is sufficient, 
ABC results in inference on the partial posterior $p(\theta|s^*)$, rather than the desired full posterior $p(\theta|y^*)$. 
Therefore, a poor choice of a summary statistic can lead to an additional bias that can be difficult to quantify and the selection of summary statistics is a crucial step in ABC \cite{Joyce_Marjoram_08, Robert11}.
A number of methods have been proposed for constructing informative summary statistics by appropriate transformations of a set of candidate summary statistics: a minimum-entropy approach \cite{Nunes2010}, regression from a set of candidate summary statistics to parameters \cite{Fearnhead2012}, or a partial least-squares transformation with boosting \cite{Aeschbacher12, drovandi2015}. A review of these methods is given in \cite{Aeschbacher12}. However, the obtained summary statistics are optimal only with respect to a given loss function (e.g., \cite{Fearnhead2012} focuses on the quadratic loss) and may not suffice for full posterior inference. In addition, they still heavily depend on the initial choice of candidate summary statistics.

Another line of research, inspired by indirect inference framework, constructs the summary statistics using auxiliary models \cite{RSSC:RSSC747, gleim2013approximate}. Here, the estimated parameters of an auxiliary model become the summary statistics for ABC. A thorough review of these method is given in \cite{drovandi2015}. The biggest advantage of this framework is that the dimensionality of the summary statistic can be determined by controlling the complexity of the auxiliary model using standard model selection techniques such as Bayesian information criterion (BIC) and Akaike information criterion (AIC). However, even if the complexity can be set in a principled way, one still needs to select the right parametric form of the auxiliary model. Furthermore, unless one uses an exponential family model, the dimensionality of the sufficient statistic will increase with the sample size according to the classical {\it{Pitman-Koopman-Darmois}} theorem (cf. e.g. \cite{Barankin1963}). Thus, for many complex models, the minimal sufficient statistic is effectively the full dataset.   


In this light, we introduce a method that circumvents an explicit selection of summary statistics.
Our method proceeds by applying a similarity measure to {\it{data themselves}}, via
embedding \cite{BerTho04, Smola2007} of the empirical data distributions into an infinite-dimensional
reproducing kernel Hilbert space (RKHS), corresponding to a positive definite kernel function defined on the data space. For suitably chosen kernels called characteristic \cite{Sriperumbudur2011}, these embeddings
capture all possible differences between distributions, e.g. all the
high-order moment information that may be missed with a set of simple summary statistics. Thus, no
information loss is incurred when going from the posterior given data $p(\theta|y^*)$
to that given the embedding $\mu(y^*)$ of data, $p(\theta|\mu(y^*))$.
A flexible representation of probability measures given by kernel embeddings has already been applied to nonparametric hypothesis testing \cite{Gretton2012}, inference in graphical models \cite{Song2013} and to proposal construction in adaptive MCMC \cite{SejStrGarAndGre14}. The key quantity arising from this framework is an easily computable notion of a distance between probability measures, termed Maximum Mean Discrepancy (MMD). When the kernel used is characteristic, a property satisfied by most commonly used kernels including Gaussian, Laplacian and inverse multiquadrics, embeddings are injective, meaning that the MMD gives a metric on probability measures. 
Here, we adopt MMD in the context of ABC as a nonparametric distance between empirical distributions of simulated and observed data. As such, there is no need to select a summary statistic first and the kernel embedding itself plays a role of a summary statistic.
In addition to the positive definite kernel used for obtaining the estimates of MMD, we apply an additional Gaussian smoothing kernel which operates on the corresponding RKHS, i.e., on embeddings themselves, to obtain a measure of similarity between simulated and observed data. 
For this reason, we refer to our method as {\it{double-kernel}} ABC, or {\it{K2-ABC}}. Our experimental results in \secref{Experiments} demonstrate that this approach results in an effective and robust ABC method which requires no hand-crafted summary statistics.

The rest of the paper is organised as follows. In \secref{background}, we overview classical approaches (rejection and soft ABC) as well as several relevant recent techniques (synthetic likelihood ABC, semi-automatic ABC, indirect score ABC, and kernel ABC) to which we will compare our method in \secref{Experiments}. 
In \secref{k2-abc}, we introduce our proposed algorithm. Experimental results
including comparisons with methods discussed in \secref{background}, 
are presented in \secref{Experiments}. We explore computational tractability of K2-ABC in  \secref{complexity}.

\section{Background}\label{sec:background}

We start by introducing ABC and reviewing existing algorithms. 
Consider a situation where it is possible to simulate a generative model and thus sample from the conditional density $p(y|\theta)$, given a value $\theta\in\Theta$
of parameters, while the computation of the likelihood $p(y^{*}|\theta)$
for the observed data $y^{*}$ is intractable. Neither exact posterior inference nor posterior sampling are possible in this case, as the posterior $p(\theta|y^{*})\propto\pi(\theta)p(y^{*}|\theta)$, for a prior $\pi$, cannot be computed up to a normalizing
constant. ABC uses an approximation of the likelihood obtained from simulation. 

The simplest form of ABC is rejection ABC. 
Let  $\epsilon>0$ be a
\emph{similarity threshold}, and $\rho$ be a notion of distance, e.g., a
premetric on domain $\mathcal{Y}$ of observations. 
The rejection ABC proceeds by sampling multiple model parameters $\theta \sim \pi$. For each 
$\theta$, a pseudo dataset $y$ is generated from $p(y|\theta)$. The parameters $\theta$ 
for which the generated $y$ are similar to the observed $y^*$, as decided by $\rho(y, y^*) <
\epsilon$,  are accepted. 
The result  is an exact sample
$\left\{ \theta_{i}\right\} _{i=1}^{M}$ from the approximated posterior
$\tilde{p}_{\epsilon}(\theta|y^{*}) \propto
\pi(\theta)\tilde{p}_{\epsilon}(y^{*}|\theta)$, where
$\tilde{p}_{\epsilon}(y^{*}|\theta) = \int_{B_{\epsilon}(y^{*})}p(y|\theta)dy$
and $B_{\epsilon}\left(y^{*}\right)  =  \left\{
    y\;:\;\rho(y,y^{*})<\epsilon\right\}$.
Choice of $\rho$ is crucial for the design of an accurate ABC algorithm.
Applying a distance directly on dataset $y$ is often challenging, when the
dataset consists of a large number of (possibly multivariate) observations.

Thus, one resorts to first choosing a summary statistic $s(y)$ and comparing them between the datasets, i.e. $\rho\left(y,y'\right)=\left\Vert s(y)-s(y')\right\Vert $. 
Since
it is generally difficult to construct sufficient statistics for complex
models, this will often ``leak'' information, e.g., if $s(y)$ represents first
few empirical moments of dataset $y$. It is only when the summary statistic $s$
is sufficient, that this approximation is consistent as $\epsilon\rightarrow0$,
i.e. that the ABC posterior $\tilde{p}_{\epsilon}(\theta|y^{*})$ will converge
to the full posterior.
Otherwise, ABC operates on the partial posterior $p(\theta|s(y^*))$, rather than the full posterior $p(\theta|y^*)$.

%
Another interpretation of the approximate likelihood $\tilde{p}_{\epsilon}(y^{*}|\theta)$
is as the convolution of the true likelihood $p(y|\theta)$ and the
``similarity'' kernel
$\kappa_{\epsilon}(y,y^{*})=\mathbf{1}\left(y\in B_{\epsilon}(y^{*})\right)$. 
Being the indicator of the $\epsilon$-ball $B_{\epsilon}(y^{*})$ computed w.r.t.\ $\rho$,
this kernel imposes a hard-constraint leading to the rejection sampling.
In fact, one can use \emph{any } similarity kernel parametrised
by $\epsilon$ which approaches delta function $\delta_{y^{*}}$
as $\epsilon\rightarrow0$.
A frequently used similarity kernel takes the form
\begin{equation}
\kappa_{\epsilon}(y,y')=\exp\left(-\frac{\rho^{q}(y,y')}{\epsilon}\right),\; \textrm{for } q>0.
\label{eq:exp_similarity_kernel}
\end{equation}
Such construction would result in a weighted sample $\left\{ \left(\theta_{j},w_{j}\right)\right\} _{j=1}^{M}$,
where $w_{j}=\frac{\kappa_{\epsilon}(y_{j},y^{*})}{\sum_{l=1}^M\kappa_{\epsilon}(y_{l},y^{*})}$, which can be directly utilised in estimating posterior expectations. 
That is, for a test function $f$, the expectation
$\int_{\Theta}f(\theta)p(\theta|y^{*})d\theta$ is estimated using
$\hat{\mathbb E} [f(\theta)]=\sum_{i=1}^Mw_j f(\theta_j)$.
This is an instance of a soft ABC, where parameter samples from the prior are weighted, rather than accepted or rejected.
\paragraph{Synthetic likelihood ABC (SL-ABC)} 

Introduced in \cite{Wood:2010aa}, the synthetic likelihood ABC models
simulated data in terms of their summary statistics and further assumes the
summary statistics have multivariate normal distribution, $s \sim \Nrm(\hat{\mu}_\theta, \hat{\Sigma}_\theta)$, with the empirical
mean and covariance defined by 
\begin{align*}
\hat{\mu}_\theta = \tfrac{1}{M} \sum_{i=1}^M s_i, \; \hat{\Sigma}_\theta = \tfrac{1}{M-1} \sum_{i=1}^M (s_i - \hat{\mu}_\theta)
(s_i - \hat{\mu}_\theta) \trp,
\end{align*} 
where $s_i$ denotes the vector of summary statistics of the $i$th simulated dataset.
Using the following similarity kernel to measure the distance from the summary statistics of actual observations $s^*$, 
$\kappa_{\epsilon}(s^*,s) = {|2\pi \epsilon I|}^{-\tfrac{1}{2}} \exp\left(-\tfrac{\left\Vert s-s^*\right\Vert_2^2}{2\epsilon^2}\right),$
the resulting synthetic likelihood is given by 
\begin{align*}
p(s^*|\theta) &= \int \kappa_{\epsilon}(s^*,s) \Nrm(s|\hat{\mu}_\theta,
\hat{\Sigma}_\theta) \; d s  \\
& = \Nrm(s^*| \hat{\mu}_\theta,
\hat{\Sigma}_\theta + \epsilon^2 I). 
\end{align*}
Relying on the synthetic likelihood, SL-ABC algorithm performs MCMC
sampling based on Metropolis-Hastings accept/reject steps with the acceptance
probability given by
%
$\alpha(\theta'|\theta) = \mbox{min}\left[
1,\frac{\pi(\theta')p(s^*|\theta')q(\theta|\theta')}{\pi(\theta)
p(s^*|\theta)q(\theta'|\theta)} \right]$,
%
where $q(\theta|\theta')$ is a proposal distribution.

\paragraph{Semi-Automatic ABC (SA-ABC)} 
Under a
quadratic loss, \cite{Fearnhead2012} shows that the optimal choice of the summary statistics is the true posterior
means of the parameters $\mathbb E\left[\theta|y\right]$ -- however, these cannot be calculated analytically. Thus,
\cite{Fearnhead2012} proposed to use the simulated data in order to construct
new summary statistics -- estimates of the posterior means of the parameters --
by fitting a vector-valued regression model from a set of candidate summary
statistics to parameters. Namely, a linear model $\theta=\mathbb
E\left[\theta|y\right]+\varepsilon=B g(y)+\varepsilon $  with the vector of
candidate summary statistics $g(y)$ used as explanatory variables (these can be
simply $g(y)=y$ or also include nonlinear functions of $y$) is fitted based on
the simulated data $\{(y_i, \theta_i)\}_{i=1}^M$. Here, assuming
$\theta\in\Theta\subset \mathbb R^d$ and $g(y)\in\mathbb R^r$, $B$ is a
$d\times r$ matrix of coefficients. The resulting estimates $s(y)=\hat B g(y)$
of $\mathbb E\left[\theta|y\right]$ can then be used as summary statistics in
standard ABC by setting $\rho(y,y')=\left\Vert s(y)-s(y')\right\Vert_2$.

\paragraph{Indirect score ABC (IS-ABC)} 
Based on an auxiliary model
$p_A(y|\phi)$ with a vector of parameters $\phi =[ \phi_1, \cdots,
\phi_{\text{dim}(\phi)}]\trp$, the indirect score ABC uses a score vector
$S_A$ as the summary statistic  \cite{gleim2013approximate}:
$S_A(y, \phi) = \left[ \tfrac{\partial \log p_A(y|\phi)}{\partial \phi_1}, \;
\cdots, \; \tfrac{\partial \log p_A(y|\phi)}{\partial \phi_{\text{dim}(\phi)}}
\right]\trp.$ 
%
When the auxiliary model parameters $\phi$ are set by the
maximum-likelihood estimate (MLE) fitted with observed data $y^*$,  the score
for the observed data $S_A(y^*, \phi_{\text{MLE}}(y^*))$  becomes zero.
Based on this fact, IS-ABC searches for the parameter values whose corresponding
simulated data produce a score close to zero. The discrepancy  between observed
and simulated data distributions in IS-ABC is given
by  
\begin{align*}
& \rho(s(y), s(y^*)) \\ 
=& \sqrt{S_A(y, \phi_{\text{MLE}}(y^*))\trp
J(\phi_{\text{MLE}}(y^*)) S_A(y, \phi_{\text{MLE}}(y^*))},
\end{align*} 
where $J(\phi_{\text{MLE}}(y^*))$ is the approximate covariance matrix of the observed score.

\paragraph{Kernel ABC (K-ABC)  } 
The use of a positive definite kernel in ABC has been explored recently in \cite{nakagome13} (K-ABC) in the context of population genetics.
In K-ABC, ABC is cast as a problem of estimating a conditional mean embedding operator mapping from summary statistics $s(y)$ to corresponding parameters $\theta$. 
The problem is equivalent to learning a regression function in the RKHSs of $s(y)$ and $\theta$ induced by their respective kernels \cite{Grunewalder2012}.
The training set $\mathcal{T}$ needed for learning the regression function is generated by firstly sampling $\{(y_i, \theta_i)\}_{i=1}^M \sim p(y|\theta)\pi(\theta)$ from which $\mathcal{T} := \{(s_i, \theta_i)\}_{i=1}^M$ by summarising each pseudo dataset $y_i$ into a summary statistic $s_i$.

In effect, given a summary statistic $s^*$ corresponding to the observations
$y^*$, the learned regression function allows one to represent the embedding of
the posterior distribution in the form of a weighted sum of the canonical feature maps $\{k(\cdot,
\theta_i)\}_{i=1}^M$ where $k$ is a kernel associated with an RKHS
$\mathcal{H}_\theta$.
In particular, if we assume that $k$ is a linear kernel (as in
\cite{nakagome13}), the posterior expectation of a function $f\in
\mathcal{H}_\theta$ is given by 
\begin{align*}
\mathbb{E}[f(\theta)|s^*] &\approx \sum_{i=1}^M w_i(s^*) f(\theta_i), \\
w_i(s^*) &= \sum_{j=1}^M ((G + M \lambda I)^{-1})_{ij} k(s_j, s^*),
\end{align*}
where $G_{ij} = g(s_i, s_j)$, $g$ is a kernel on $s$, and $\lambda$ is a regularization parameter.
The use of a kernel $g$ on summary statistics $s$ implicitly transforms $s$ non-linearly, thereby increasing the representativeness of $s$. 
Nevertheless, the need for summary statistics is not eliminated. 



\section{Proposed method}\label{sec:k2-abc}
We first overview kernel MMD, a notion of distance between probability measures
that is used in the proposed K2-ABC algorithm.

\paragraph{Kernel MMD  } 
For a probability distribution $F_x$ on a domain $\mathcal X$, its kernel embedding is defined as $\mu_{F_x} = \Em_{X\sim F_x} k(\cdot, X)$ \cite{Smola2007}, an element of an RKHS $\mathcal H$ associated with a positive definite kernel $k:\mathcal X\times \mathcal X\to\mathbb R$. 
An embedding exists for any $F_x$ whenever the kernel $k$ is bounded, or if
$F_x$ satisfies a suitable moment condition w.r.t.\ an unbounded kernel $k$
\cite{SejSriGreFuk13}. For any two given probability measures $F_x$ and $F_y$,
their maximum mean discrepancy (MMD) is simply the Hilbert space distance
between their embeddings:
\begin{align*}
& \text{MMD}^2(F_x, F_y) = \left\Vert\mu_{F_x}-\mu_{F_y}
\right\Vert^2_{\mathcal{H}} \\
 =& \Em_X \Em_{X'} k(X,X')+\Em_Y \Em_{Y'} k(Y,Y') -2\Em_X \Em_Y k(X,Y),
\end{align*}
where $X,X'\iid F_x$ and $Y,Y'\iid F_y$. 
While simple kernels like polynomial of order $r$ capture differences in first $r$ moments of distributions, particularly interesting are kernels with a \emph{characteristic} property
\footnote{A related notion of \emph{universality} is often employed.}
\cite{Sriperumbudur2011}, for which the kernel embedding is
injective and thus MMD defined by such kernels gives a metric on the space of probability distributions.
Examples of such kernels include widely used kernels such as Gaussian RBF and
Laplacian.  Being written in terms of expectations of kernel functions allows
straightforward estimation of MMD on the basis of samples: given $\left\{
    x^{(i)}\right\}_{i=1}^{n_{x}}\iid F_x
,\; \left\{ y^{(j)}\right\}_{j=1}^{n_{y}}\iid F_y$, an unbiased estimator is given by
\begin{eqnarray*}
\widehat{\text{MMD}}^2(F_x,F_y)
&=& \frac{1}{n_{x}(n_{x}-1)}\sum_{i=1}^{n_{x}} \sum_{j\neq i}k(x^{(i)},x^{(j)}) \\
&& +\frac{1}{n_{y}(n_{y}-1)} \sum_{i=1}^{n_{y}}\sum_{j\neq i}k(y^{(i)},y^{(j)}) \\
&& -\frac{2}{n_{x}n_{y}}\sum_{i=1}^{n_{x}}\sum_{j=1}^{n_{y}}k(x^{(i)},y^{(j)}).
\end{eqnarray*}
Further operations are possible on kernel embeddings - one can define a
positive definite kernel on probability measures themselves using their
representation in a Hilbert space. An example of a kernel on probability measures 
is
$\kappa_\epsilon(F_x,F_y)  =
\exp\left(-\frac{\text{MMD}^{2}(F_x,F_y)}{\epsilon}\right)$,
\cite{Christmann10} with $\epsilon>0$.
 This has
recently led to a thread of research tackling the problem of learning on
distributions, e.g., \cite{szabo14}.  These insights are essential to our
contribution, as we employ such kernels on probability measures in the design
of the K2-ABC algorithm which we describe next.

\paragraph{K2-ABC}
The first component of K2-ABC is a nonparametric distance $\rho$ between
empirical data distributions.
Given two datasets $y=\left(y^{(1)},\ldots,y^{(n)}\right)$ and
$y'=\left(y'^{(1)},\ldots,y'^{(n)}\right)$ consisting of $n$ i.i.d.\
observations\footnote{The i.i.d.\ assumption can be relaxed in practice, as we
demonstrate in \secref{Experiments} on time series data.}, we use MMD to
measure the distance between $y, y'$:
$\rho^2(y,y')=\widehat{\text{MMD}}^2({F}_{y},{F}_{y'})$, i.e. $\rho^2$ is
an unbiased estimate of $\text{MMD}^2$ between probability distributions
$F_y$ and $F_{y'}$ used to generate $y$ and $y'$. 
This is almost the same as setting empirical kernel embedding $s(y)=\mu_{\hat
F_y}=\sum_{j=1}^n k\left(\cdot,y^{(j)}\right)$ to be the summary statistic. 
However, in that case $\left\Vert s(y)-s(y')\right\Vert^2_\mathcal{H} =
\text{MMD}^{2}(\hat{F}_x,\hat{F}_y)$ would have been a biased estimate of the
population $\text{MMD}^2$ \cite{Gretton2012}. 
Our choice of $\rho$ is
guaranteed to capture all possible differences (i.e. all moments) between
$F_y$ and $F_{y'}$ whenever a characteristic kernel $k$ is employed
\cite{Sriperumbudur2011}, i.e. we are operating on a full posterior and there
is no loss of information due to the use of insufficient statistics.

\begin{algorithm}[h]
\caption{K2-ABC Algorithm}
\label{algo:k2abc}
\begin{algorithmic}
\REQUIRE observed data $y^*$, prior $\pi$, soft threshold $\epsilon$
\ENSURE Empirical posterior $\sum_{i=1}^M w_i \delta_{\theta_i}$
\FOR{$i=1,\ldots,M$}
   \STATE Sample $\theta_i\sim\pi$
   \STATE Sample pseudo dataset $y_i \sim p(\cdot|\theta_i)$
   \STATE $\widetilde{w}_i = \exp\left(-\frac{\widehat{\text{MMD}}^2({F}_{y_i},{F}_{y^*})}{\epsilon}\right)$
\ENDFOR
\STATE $w_i = \widetilde{w}_i / \sum_{j=1}^M \widetilde{w}_j \quad$ for $i=1,\ldots,M$
\end{algorithmic}
\end{algorithm}
Further, we introduce a second kernel into the ABC algorithm (summarised in
Algorithm~\ref{algo:k2abc}), the one that operates directly on probability
measures, and compute the ABC posterior sample weights,
\begin{equation}
\kappa_\epsilon(F_y,F_{y'})  =
\exp\left(-\frac{\widehat{\text{MMD}}^{2}(F_y,F_{y'})}{\epsilon}\right),
\end{equation}
with a suitably chosen parameter $\epsilon>0$.  Now, the datasets are
compared using the estimated similarity $\kappa_\epsilon$ between their
generating distributions. There are two sets of parameters in K2-ABC,
parameters of kernel $k$ (on original domain) and $\epsilon$ in the kernel
$\kappa_\epsilon$ (on probability measures).  

K2-ABC is readily applicable to  non-Euclidean input objects if a kernel $k$ can be 
defined on them. Arguably the most common application of ABC is to genetic data. 
Over the past years there have been a number of works addressing the use of
kernel methods for studying genetic data including \cite{Wu2010} which considered
genetic association analysis, and \cite{Li2012} which studied gene-gene interaction.
Generic kernels for strings, graphs and other structured data have also been
explored \cite{Gaertner2003}.

\section{Experiments}\label{sec:Experiments}


\paragraph{Toy problem} 

\begin{figure*}[t] 
\centering
\includegraphics[width=0.95\textwidth]{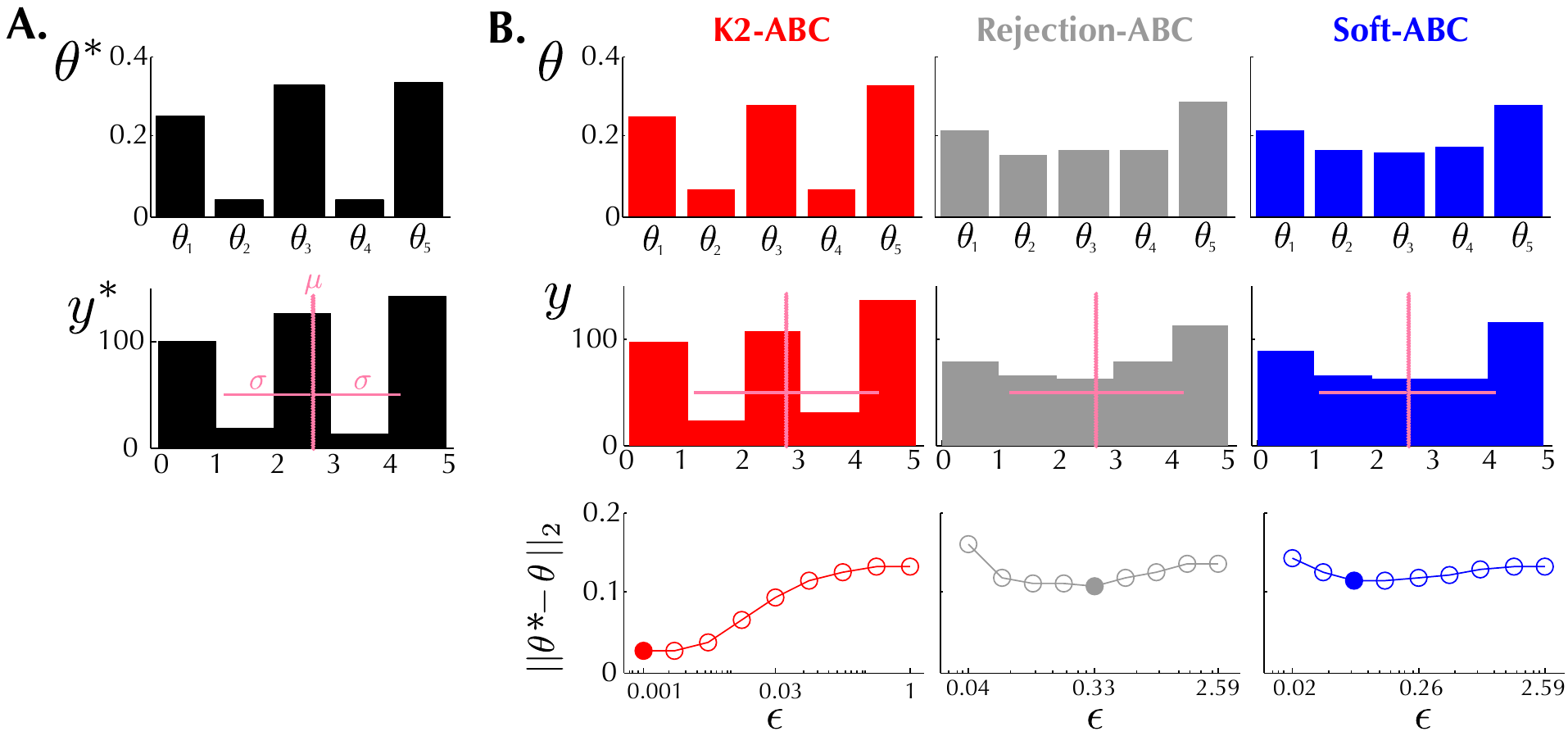}
\caption{
A possible scenario for ABC algorithms to fail due to insufficient summary statistics. \textbf{A}: Using the 5-dimensional true parameters (top), $400$ observations (bottom) are sampled from the mixture of uniform distributions in \eqref{likelihood_mixture_uniform}.  
\textbf{B (top)}:  Estimated posterior mean of parameters from each method. 
We drew $1000$ parameters from the Dirichlet prior and $400$ simulated data points given each parameter. 
In rejection and soft ABC algorithms, we used empirical mean and variance of observations as summary statistics to determine similarity between simulated and observed data. 
\textbf{B (middle)}:  Histograms of $400$ simulated data points given estimated posterior means by each method. 
Though the mean and variance of simulated data from rejection and soft ABC match that of the observed data, the shapes of the empirical distributions notably differ.
\textbf{B (bottom)}: Euclidean distance between true and estimated posterior mean of parameters as a function of $\epsilon$.  
We varied the $\epsilon$ values to find the optimal range in terms of the Euclidean distance. 
The magnitude of $\epsilon$ is algorithm-specific and not comparable across methods.
}
\label{fig:toy_mixture_uniform}
\end{figure*}

We start by illustrating how the choice of summary statistics can significantly affect the inference result, especially when the summary statistics are not sufficient. 
We consider a symmetric Dirichlet prior $\pi$ and a likelihood $p(y|\theta)$ given by a mixture of uniform distributions as
\begin{align}
    \pi(\theta) &= \text{Dirichlet}(\theta; \boldsymbol{1}), \nonumber \\
    p(y|\theta) &= \sum_{i=1}^5 \theta_i \text{Uniform}(y; [i-1, i]).
    \label{eq:likelihood_mixture_uniform}
\end{align}
The model parameters $\theta$ are a vector of mixing proportions.
The goal is to estimate $\mathbb{E}[\theta|y^*]$ where $y^*$ is generated with true parameter $\theta^* = [0.25, 0.04, 0.33, 0.04, 0.34]^\top$ (see \figref{toy_mixture_uniform}A).
The summary statistics are chosen to be empirical mean and variance i.e. $s(y) = (\hat{\mathbb{E}}[y], \hat{\mathbb{V}}[y] )^\top$.

We compare three ABC algorithms: K2-ABC, rejection ABC, and soft ABC.
Here, soft ABC refers to an ABC algorithm which uses a similarity kernel in \eqref{exp_similarity_kernel} with $q=2$ and 
$\rho(y, y') = \|s(y) - s(y') \|_2$.
For K2-ABC, a Gaussian kernel defined as $k(a, b) = \exp\left( -\frac{\|a - b\|_2^2}{2\gamma^2} \right)$
%
%
is used where $\gamma$ is set to $\text{median}(\{\|y^{*(i)} - y^{*(j)} \| \}_{i,j})$ \cite{Scholkopf2001}. 
We test different values of $\epsilon$ on a coarse grid, and report the estimated $\mathbb{E}[\theta|y^*]$  which is closest to $\theta^*$ as measured with a Euclidean distance.

The results are shown in \figref{toy_mixture_uniform} where the top row shows the estimated $\mathbb{E}[\theta | y^*]$ from each method, associated with the best $\epsilon$ as reported in the third row.
The second row of \figref{toy_mixture_uniform}, from left to right, shows $y^*$ and $400$ realizations of $y$ drawn from $p(y|\mathbb{E}[\theta|y^*])$ obtained from the three algorithms.
In all cases, the mean and variance of the drawn realizations match that of $y^*$.
However, since the first two moments are insufficient to characterise $p(y|\theta^*)$, there exists other $\theta'$ that can give rise to the same $s(y')$, which yields inaccurate posterior means shown in the top row. 
In contrast, K2-ABC taking into account infinite-dimensional sufficient statistic correctly estimates the posterior mean.


\begin{figure*}[t] 
\centerline{\includegraphics[width=0.9\textwidth]{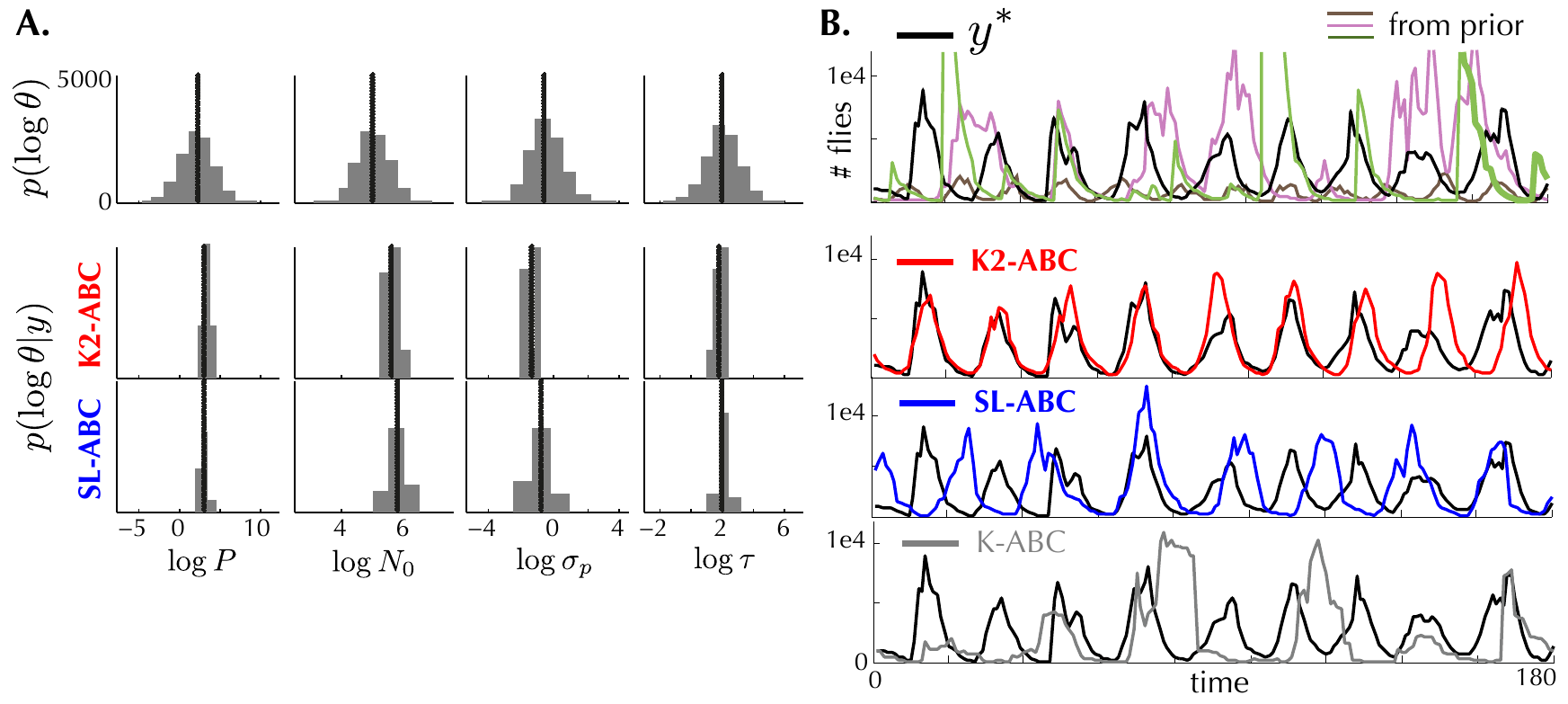}}
\caption{Blowfly data.  
\textbf{A (top)}: Histograms of $10,000$ samples for four parameters $\{ \log P, \log N_0, \log \sigma_p, \log \tau \}$ drawn from the prior. 
\textbf{A (middle/bottom)}: Histogram of samples from the posteriors obtained by K2-ABC / SL-ABC (acceptance rate: $0.2$, burn-in: $5000$ iterations), respectively. 
In both cases, the posteriors over parameters are concentrated around their means (black bar).  
The posterior means of $P$ and $\tau$ obtained from K2-ABC are close to those obtained from SL-ABC, while there is noticeable difference in the means of $N_0$ and $\sigma_p$.  
Note that we were not able to show the same histogram for K-ABC since the posterior obtained by K-ABC is improper.
\textbf{B (top)}: Three realisations of $y$ given three different parameters drawn from the prior. 
Small changes in $\theta$ drastically change $y$.  \textbf{B (middle to bottom)}: Simulated data using inferred parameters (posterior means) shown in A. Our method (in red) produces the most similar dynamic trajectory to the actual observation (in black) among all the methods.}
\label{fig:Figure2}
\end{figure*}

\paragraph{Ecological dynamic systems } 
As an example of statistical inference for ecological dynamic systems, we use observations on adult blowfly populations over time introduced in \cite{Wood:2010aa}. 
The population dynamics are modelled by a discretised differential equation: 
\begin{align*}
N_{t+1} &= P N_{t-\tau} \exp\left(-\frac{N_{t-\tau}}{N_0} \right)e_t + N_t \exp(-\delta \epsilon_t),
\end{align*} 
where an observation at time $t+1$ is denoted by $N_{t+1}$ which is determined by time-lagged observations $N_t$ and $N_{t-\tau}$ as well as Gamma distributed noise realisations $e_t \sim \mbox{Gam}(\frac{1}{\sigma_p^2}, \sigma_p^2)$ and $\epsilon_t \sim \mbox{Gam}(\frac{1}{\sigma_d^2}, \sigma_d^2)$. 
Here, the parameters are $\theta = \{P, N_0,  \sigma_d, \sigma_p, \tau, \delta \} $. 
We put broad Gaussian priors on log of parameters as shown in \figref{Figure2}A.  
Note that the time series data given the parameters drawn from the priors vary drastically (see \figref{Figure2}B), and therefore inference with those data is very challenging as noted in \cite{Meeds14}. 

The observation (black trace in \figref{Figure2}B) is a time-series of length $T = 180$, where each point in time indicates how many flies survive at each time under food limitation. For SL-ABC and K-ABC, we adopted the custom 10 summary statistics used in \cite{Meeds14}: the log of the mean of all $25\%$ quantiles of $\{ N_t/1000\}_{t=1}^T$ (four statistics), the mean of  $25\%$ quantiles of the first-order differences of $\{ N_t/1000\}_{t=1}^T$ (four statistics), and the maximal peaks of smoothed $\{ N_t\}_{t=1}^T$, with two different thresholds (two statistics).  For IS-ABC, following \cite{drovandi2015}, we use a Gaussian mixture model with three components as an auxiliary model. In addition, we ran two versions of SA-ABC algorithm on this example: SAQ regresses to simulated parameters from the corresponding simulated instances of time-series appended with the quadratic terms , i.e., $g(y)=(y,y^2)\in\mathbb R^{2T}$, whereas SA-custom uses the above custom 10 summary statistics from \cite{Meeds14} appended with their squares as the candidate summary-statistics vector $g(y)$ (which is thus 20-dimensional in this instance).\footnote{For SL-ABC, we used the Python implementation by the author of \cite{Meeds14}. For IS-ABC, we used the MATLAB implementation by the author of \cite{drovandi2015}. For SA-ABC, we used the R package \texttt{abctools} written by the author of \cite{Fearnhead2012}.}

For setting $\epsilon$ and kernel parameters in K2-ABC, we split the data into
two sets: training ($75\%$ of $180$ data points) and test (the rest) sets. 
Using the training data, we ran each ABC algorithm given each value of
$\epsilon$ and kernel parameters defined on a coarse grid, then, computed test
error\footnote{We used Euclidean distance between the histogram (with 10 bins)
of test data and that of predictions made by each method. We chose the
difference in histogram rather than in the realisation of $y$ itself, to avoid
the error due to the time shift in $y$.} to choose the optimal values of
$\epsilon$ and kernel parameters in terms of the minimum prediction error.
Finally, with the chosen $\epsilon$ and kernel parameters, we ran each ABC
algorithm using the entire data.

\begin{figure}[h]
\centering
\includegraphics[width=1\columnwidth]{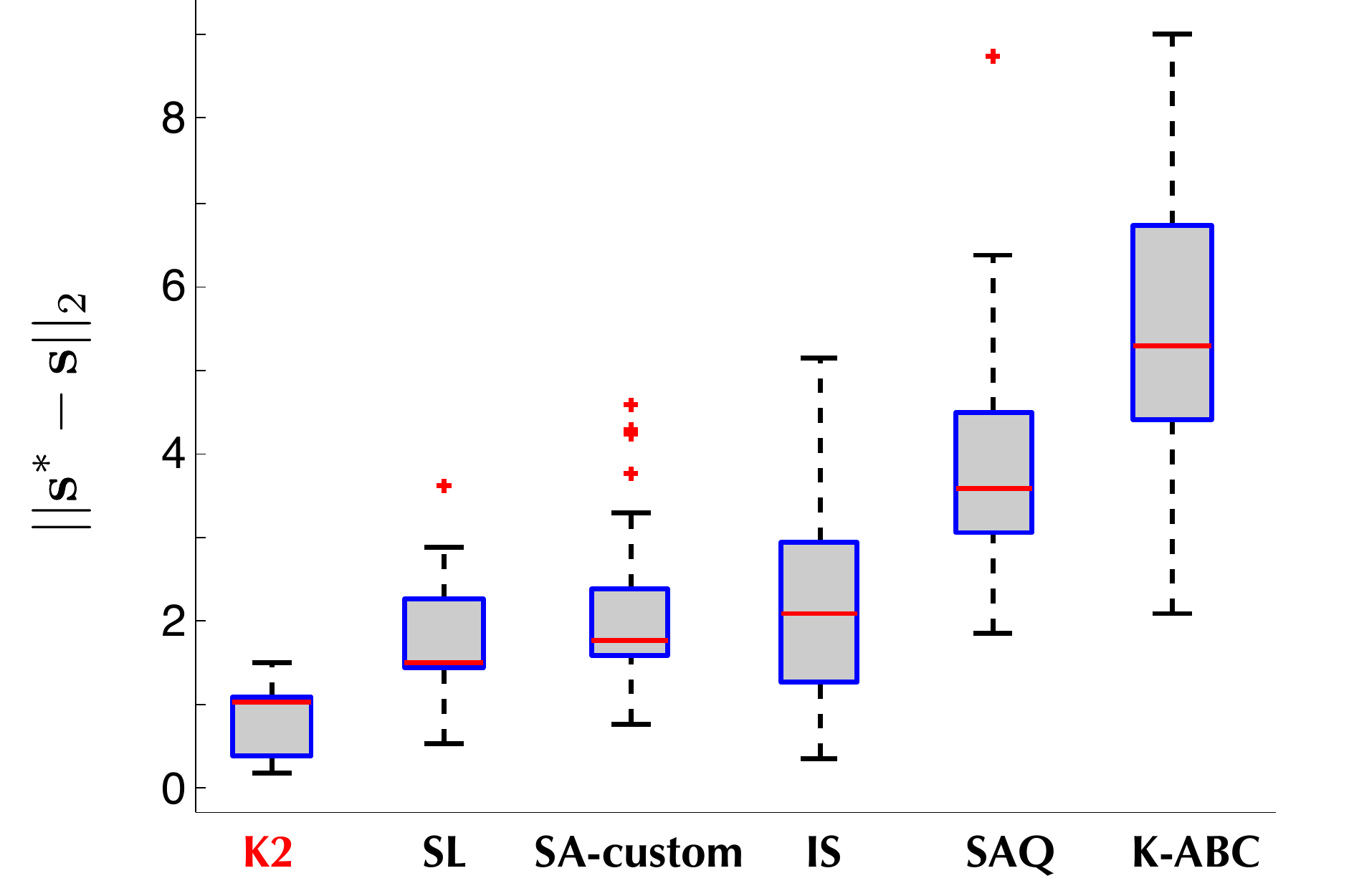}
\caption{Euclidean distance between the chosen $10$ summary statistics of $y^*$
and $y$ given the posterior mean of parameters estimated by various methods.
Due to the fluctuations in $y$ from the noise realisations ($\epsilon_t, e_t$),
we made $100$ independent draws of $y$ and  computed the Euclidean distance.
Note that K2-ABC achieved nearly $50\%$ lower errors than the next best method,
SL-ABC, although SL-ABC, SA-custom, and K-ABC all explicitly operate on the
summary statistics in the comparison while K2-ABC does not.}
\label{fig:Figure3}
\end{figure}

We show the concentrated posterior mass after performing K2-ABC in \figref{Figure2}A, as well as
an example trajectory drawn from the inferred posterior mean  in \figref{Figure2}B \footnote{Note that reproducing the trajectory exactly {\it{is not}} the main goal of this experiment. We show the example trajectory to give the readers a sense of what the trajectory looks like.}. To quantify the accuracy of each method, we compute the Euclidean
distance between the vector of chosen $10$ summary statistics $s^*=s(y^*)$ for the observed data and $s(y)$ for the simulated data $y$ given the estimated posterior mean $\hat\theta$ of the parameters. As shown in \figref{Figure3}, K2-ABC outperforms other methods, although SL-ABC, SA-custom, and K-ABC all explicitly operate on this vector of summary statistics $s$ while K2-ABC does not. In other words, while those methods attempt to explicitly pin down the  parts of the parameter space which produce summary statistics $s$ similar to $s^*$, insufficiency of these summary statistics affects the posterior mean estimates undesirably even with respect to that very metric.


\section{Computational tractability}\label{sec:complexity}

In K2-ABC, given a dataset and pseudo dataset with $n$ observations each, the cost for computing
$\widehat{\mathrm{MMD}}^2(F_{y_i}, F_{y^*})$ is $O(n^2)$. For $M$ pseudo datasets, the total cost 
 then becomes $O(Mn^2)$, which can be prohibitive for a large number of observations. 
Since computational tractability is among the core considerations for ABC, in this section, we examine the performance of K2-ABC with different MMD approximations which reduce the computational cost. 

\paragraph{Linear-time MMD}
 
The unbiased linear-time MMD estimator presented in \cite[section 6]{Gretton2012} reduces the total cost to $O(Mn)$ at the price of a higher variance. 
Due to its computational advantage, the linear-time MMD has been successfully applied in large-scale two-sample testing \cite{Gretton2012b} as a test statistic. The original linear-time MMD is given by 
\begin{align*}
&\widehat{\mathrm{MMD}}_{l}^{2}(F_{x},F_{y}) \\
&= \frac{2}{n}\sum_{i=1}^{n/2}\bigg[k(x^{(2i-1)},x^{(2i)})
 +k(y^{(2i-1)},y^{(2i)}) \\
& \quad -k(x^{(2i-1)},y^{(2i)})-k(x^{(2i)},y^{(2i-1)})\bigg].
\end{align*}
Note that we have assumed the same number of observations $n_x=n_y=n$ from $F_x$ and $F_y$. The estimator $\widehat{\mathrm{MMD}}_{l}^{2}$ is constructed so that the independence of the summands allows 
derivation of its asymptotic distribution and the corresponding quantile computation needed for two-sample testing. However, since we do not require such independence, we employ a linear-time estimator with a larger number of summands, which also does not require $n_x=n_y$.
Without loss of generality, we assume $n_{x}\leq n_{y}$ 
Denote $x^{(j)}:=x^{(1+\mathrm{mod}(j-1,n_{x}))}$
for $j>n_x$, i.e., we allow a cyclic shift through the smaller dataset $\{x^{(i)}\}_{i=1}^{n_{x}}$. The linear-time
MMD estimator that we propose is 
{\small
\begin{align*}
& \widehat{\mathrm{MMD}}_{L}^{2}(F_{x},F_{y})=
\frac{1}{n_{x}-1}\sum_{i=1}^{n_{x}-1}k(x^{(i)},x^{(i+1)}) \\
& +\frac{1}{n_{y}-1}\sum_{i=1}^{n_{y}-1}k(y^{(i)},y^{(i+1)})
-\frac{2}{n_{y}}\sum_{i=1}^{n_{y}}k(x^{(i)},y^{(i)}),
\end{align*}}%
which is an unbiased estimator of $\mathrm{MMD}^2(F_x, F_y)$. The total cost of the resulting K2-ABC algorithm is $O(M(n_x+ n_y))$.

\paragraph{MMD with random Fourier features}

Another fast linear MMD estimator can be achieved by considering an
approximation to the kernel function $k(x,y)$ with an inner product
of finite dimensional feature vectors $\hat{\phi}(x)^{\top}\hat{\phi}(y)$
where $\hat{\phi}(x)\in\mathbb{R}^{D}$ and $D$ is the number of
features. Given the feature map $\hat{\phi}(\cdot)$ such that,
$k(x,y)\approx\hat{\phi}(x)^{\top}\hat{\phi}(y)$,
MMD$^2$ can be approximated as
{\small
\begin{align*}
& \mathrm{MMD}_{rf}^{2}(F_{x},F_{y})   \\
%
& \approx \mathbb{E}_{X}\hat{\phi}(X)^{\top}\mathbb{E}_{X'}
  \hat{\phi}(X')+\mathbb{E}_{Y}\hat{\phi}(Y)^{\top}\mathbb{E}_{Y'}
  \hat{\phi}(Y') \\
  & \quad -2\mathbb{E}_{X}\hat{\phi}(X)^{\top}\mathbb{E}_{Y}\hat{\phi}(Y)\\
& :=\|\mathbb{E}_{X}\hat{\phi}(X)-\mathbb{E}_{Y}\hat{\phi}(Y)\|_{2}^{2}.
\end{align*}
}%
A straightforward (biased) estimator is 
\[
\widehat{\mathrm{MMD}}_{rf}^{2}(F_{x},F_{y})=\bigg\|\frac{1}{n_{x}}\sum_{i=1}^{n_{x}}\hat{\phi}(x^{(i)})-\frac{1}{n_{y}}\sum_{i=1}^{n_{y}}\hat{\phi}(y^{(i)})\bigg\|_{2}^{2},
\]
which can be computed in $O(D(n_{x}+n_{y})$), i.e., linear in the
sample size, leading to the overall cost of $O(MD(n_x+ n_y))$.


Given a kernel $k$, there are a number of ways to obtain $\hat{\phi}(\cdot)$
such that $k(x,y)\approx\hat{\phi}(x)^{\top}\hat{\phi}(y)$. One
approach which became popular in recent years is based on random Fourier
features \cite{Rahimi07randomfeatures} which can be applied to any
translation invariant kernel. Assume that $k$ is translation invariant
i.e., $k(x,y)=\tilde{k}(x-y)$ for some function $\tilde{k}$. According
to Bochner's theorem \cite{Rudin2013}, $\tilde{k}$ can be written
as
\begin{align*}
\tilde{k}(x-y) & =\int e^{i\omega^{\top}(x-y)}\,\mathrm{d}\Lambda(\omega)\\
& =\mathbb{E}_{\omega\sim\Lambda}\cos(\omega^{\top}(x-y)) \\
 & =2\mathbb{E}_{b\sim U[0,2\pi]}\mathbb{E}_{\omega\sim\Lambda}
 \cos(\omega^{\top}x+b)\cos(\omega^{\top}y+b),
\end{align*}
where $i=\sqrt{-1}$ and due to positive-definiteness of $\tilde k$, its Fourier transform
$\Lambda$ is nonnegative and can be treated as a probability measure. By drawing
random frequencies $\{\omega_{i}\}_{i=1}^{D}\sim\Lambda$
and $\{b_{i}\}_{i=1}^{D}\sim U[0,2\pi]$, $\tilde{k}(x-y)$ can be
approximated with a Monte Carlo average. It follows that $\hat{\phi}_{j}(x)=\sqrt{2/D}\cos(\omega_{j}^{\top}x+b_{j})$
and $\hat{\phi}(x)=(\hat{\phi}_{1}(x),\ldots,\hat{\phi}_{D}(x))^{\top}$.
Note that a Gaussian kernel $k$ corresponds to
normal distribution $\Lambda$.

\paragraph{Empirical results} 
We employ the linear-time and the random Fourier feature MMD estimators in our K2-ABC algorithm, which we call {\it{K2-lin}} and {\it{K2-rf}}, respectively, and test these variants on the blowfly data. For K2-rf, we used $50$ random features. 
\begin{SCfigure}[30][h]
\centering
\includegraphics[width=0.6\columnwidth]{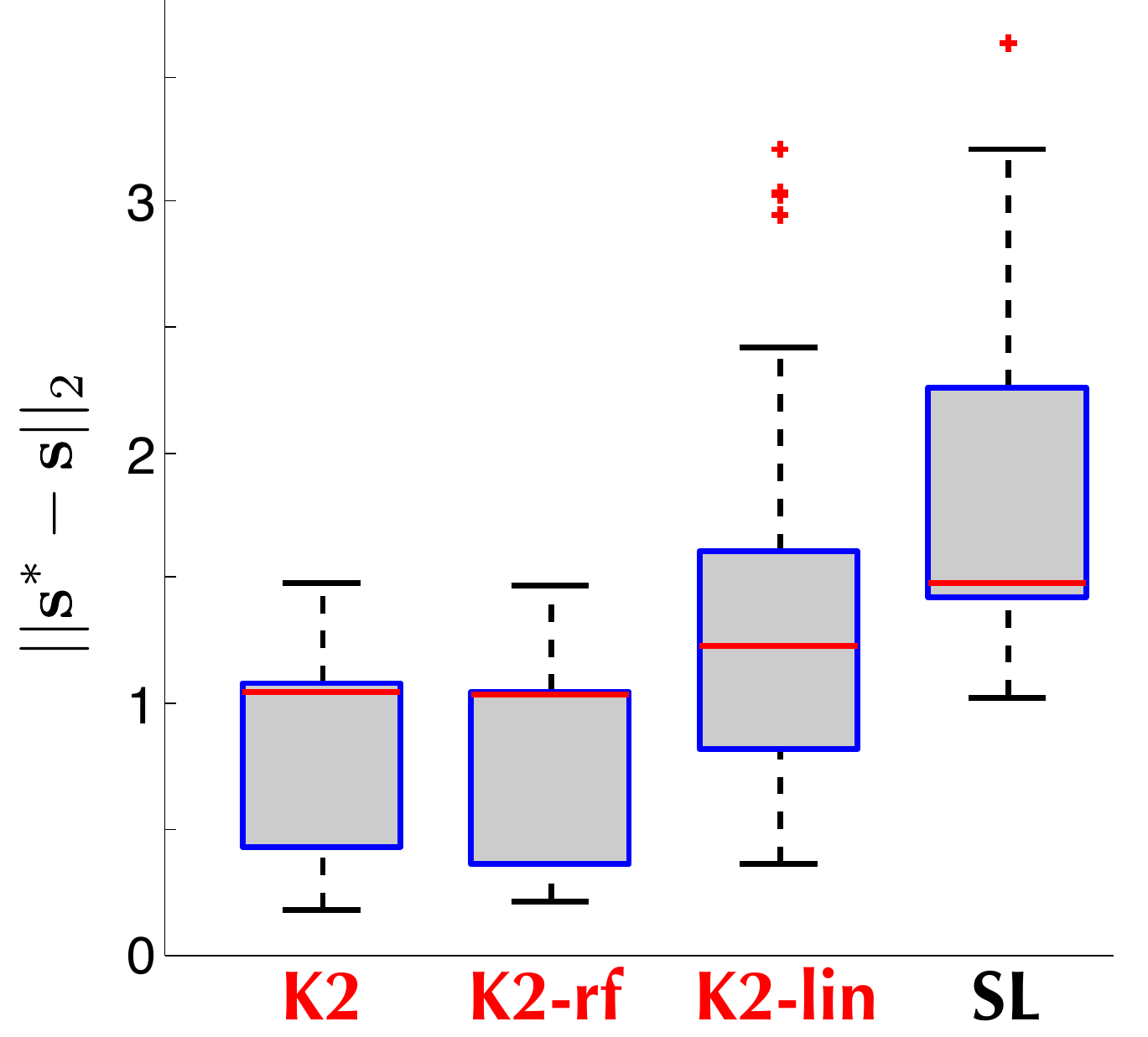}
\caption{K2-ABC with different MMD estimators outperform the best existing
method, SL-ABC, on the blowfly data.}
\label{fig:Figure4}
\end{SCfigure}

\section{Conclusions and further work}

We investigated the feasibility of using MMD as a discrepancy measure of 
samples from two distributions in the context of ABC. 
Via embeddings of empirical data distributions into an RKHS, we effectively take into account infinitely many implicit features of these distributions as summary statistics. When tested on both simulated and real-world datasets,  our approach obtained more accurate posteriors, compared to other methods that rely on hand-crafted summary statistics. 

While any choice of a characteristic kernel will guarantee infinitely many features and no information loss due to the use of partial posteriors, we note that the kernel choice is nonetheless important for MMD estimation and therefore also for the efficiency of the proposed K2-ABC algorithm. As widely studied in the RKHS literature, the choice should be made to best capture characteristics of given data, i.e., by utilising domain-specific knowledge. For instance, when some data components are believed {\it{a priori}} to be on different scales, one can adopt the automatic relevance detemination (ARD) kernel instead of the Gaussian kernel. Formulating explicit efficiency criteria in the context of ABC and optimizing over kernel choice, similarly as in the context of two-sample testing \cite{Gretton2012b}, would be an essential extension. 



\bibliographystyle{unsrt}   
\bibliography{abc_biblio}



%


\end{document}